\documentclass[a4paper]{llncs}
\pdfoutput=1
\usepackage[english]{babel}
\usepackage[T1]{fontenc}
\usepackage[utf8]{inputenc}
\usepackage{lipsum}
\usepackage[nocompress]{cite}
\usepackage{graphicx}
\usepackage{color}
\usepackage{amsmath}
\usepackage{amssymb}
\usepackage{todonotes}
\usepackage{hyperref}
\usepackage{url}
\usepackage{rotating}
\usepackage{wrapfig}
\usepackage{soul}
\usepackage{xr}
\usepackage{tikz}
\usepackage{pgfplots}
\usetikzlibrary{shapes,arrows}
\usepackage{amsmath}

\presetkeys{todonotes}{
  size=\tiny,
}{}

\newsavebox\quotequationbox
\newenvironment{quotequation}
  {\begin{equation}
   \begin{lrbox}{\quotequationbox}
   \begin{minipage}{\dimexpr\columnwidth-2\leftmargini}
   \setlength{\leftmargini}{0pt}%
   \begin{quote}}
  {\end{quote}
   \end{minipage}
   \end{lrbox}\makebox[0pt]{\usebox{\quotequationbox}}
   \end{equation}}

\title{Using Recurrent Neural Network for Learning Expressive Ontologies}
\author{Giulio Petrucci\inst{1,2}, Chiara Ghidini\inst{1}, Marco Rospocher\inst{1}} \institute{FBK-irst, Via Sommarive, 18, I-38123 Trento, Italy\\ \and University of
Trento, Via Sommarive, 14, I-38123 Trento, Italy\\
{\fontsize{8}{10}\selectfont\email{\{petrucci,ghidini,rospocher\}@fbk.eu}}
}

\begin{document}
\maketitle

\begin{abstract}
Recently, Neural Networks have been proven extremely effective in many natural language processing tasks such as sentiment analysis, question answering, or machine translation. 
Aiming to exploit such advantages in the Ontology Learning process, in this technical report we present a detailed description of a Recurrent Neural Network based system to be used to pursue such goal. 
\end{abstract}

\section{Introduction}
\label{sec:introduction}

Large part of the knowledge to be encoded in ontologies comes from
semi-structured or more likely unstructured sources like natural
language texts, being them collection of documents or transcriptions
of interviews of domain experts. The task to manually produce and
maintain ontologies is complex and time-consuming -- specially if the
available data are continuously increasing their volume, variety and
velocity of production, like in the case of the World Wide Web.
Complexity and time-consumption increase if we set the target of the
ontology building process to a very expressive ontology, with complex
axioms capturing in deep the meaning of the portion of knowledge we
want to represent.  Many approaches, together with many software
tools, have been proposed along the years in order to reduce the
ontology engineering cost. But despite such progress, quoting authors
of \cite{volker2008lexo}, state of the art methods are \emph{able to
  generate ontologies that are largely informal}, that is
\emph{limited in their expressiveness} but not suitable to capture a
higher expressiveness at an affordable human effort.

In the last years, Neural Networks have accomplished excellent result
in many Natural Language Processing task (see
\cite{collobert2011nlp}): part-of-speech tagging, dependency parsing
(see \cite{socher2011parsing}), semantic role labelling (see
\cite{collobert2008unified}), question-answering, sentiment analysis
(see \cite{socher2013recursive}), statistical machine translation (see
\cite{hermann2013multilingual}) Some insightful approaches have been
proposed also for more complex tasks like learning logical semantics
(see \cite{bowman2015recursive}), some sort of reasoning (see
\cite{peng2015towards}) and executing small sets of instructions
encoded as a short piece of programming code (see
\cite{zaremba2014learning}).  The great advantage of using Neural
Networks is their capability to capture meaningful features in text,
being also cheap in terms of feature engineering and capable to
exploit large amount of unlabeled training data.  In particular,
\emph{Recurrent Neural Networks} have been successfully used in Natural
Language Processing as their inner recurrent structure is naturally
suitable for handling natural language, where the meaning of each word
is naturally dependent on the ones preceding and following.

Our idea is that Recurrent Neural Networks can be exploited in the
ontology learning task, as they can handle the typical syntactic
structures of encyclopaedic text, relieving the ontology engineer from
the burden of manually encode large catalogues of rules and
patterns. To achieve such goal, we designed a Recurrent Neural Network
based system endowed with some short-term memory capabilities through
Gated Recursive Units. This architecture is currently under evaluation.

The paper is structured as follows: in Section \ref{sec:taskdesc} we
give a formulation for the ontolgy learning task in order to better
understand the motivation of our approach. In Section
\ref{sec:rnnsforol} we describe in detail the system we
designed. Section \ref{sec:conclusion} concludes the paper.



\section{Ontology Learning as a Transduction task}
\label{sec:taskdesc}

Let's consider the following simple description of a bee and its main characteristics:

\begin{quotequation}
  \label{eq:firstsentence}
  \centering
  \emph{``A bee is an insect that has 6 legs and produces honey.''} 
\end{quotequation}

We can think of encoding the same knowledge in a DL-formula, like:

\begin{equation}
  \label{eq:definition}
  \mathtt{Bee} \sqsubseteq \mathtt{Insect} \sqcap =\mathtt{6} \ \mathtt{have}.\mathtt{Leg} \sqcap \exists \mathtt{produce}.\mathtt{Honey}
\end{equation}

where nouns like \texttt{Bee}, \texttt{Insect}, \texttt{Leg} and \texttt{Honey} denote the involved \emph{concepts} while verbs like \texttt{have} and \texttt{produce} describe \emph{roles}, i.e. relationships  among such concepts. The same knowledge could be expressed in a different text surface instance, like:

\begin{quotequation}
  \label{eq:altsentence}
  \centering
  \emph{``If something is a bee, then it is an insect \\ with exactly 6 legs and it also produces honey.''}
\end{quotequation}

Dually, we could consider the example of another sentence, dealing with a content which is radically different from \eqref{eq:firstsentence} but \emph{structurally} similar, as:

\begin{quotequation}
  \label{eq:diffsentence}
  \centering
  \emph{``A cow is a mammal that has 2 horns and eats grass''}
\end{quotequation}

that, translated into an OWL formula would be again \emph{structurally} similar to \eqref{eq:definition}.
Roughly speaking, we observe that sentences with similar content tend to share \emph{content words} -- nouns, adjectives, verbs -- while those with similar structure tend to share function words -- determiners, adverbs, articles, pronouns.

\begin{figure}[t]
  \centering
  \scalebox{.7}{\begin{tikzpicture}[node distance=8em]
\node [name=s] {\texttt{sentence}};
\node [name=str, draw, 
right of=s,
minimum height=4em, 
text width=7em, 
text centered] {\emph{Sentence \\ Transduction}};
\node [name=f, right of=str, xshift=2em] {\texttt{template formula}};
\node [name=st, draw, 
below of=str,
minimum height=4em, 
text width=7em, 
text centered,
yshift=1em] {\emph{Sentence \\ Tagging}};
\node [name=ts, right of=st, xshift=2em] {\texttt{tag set}};
\node [draw, name=mul, right of=f, shape=circle, dashed, xshift=-1em] {$\times$};
\node [name=af, right of=mul, xshift=-3em] {\texttt{formula}};

\path [draw, -latex'] (s.east) -- (str.west);
\path [draw, -latex'] (str.east) -- (f.west);
\path [draw, -latex'] (f.east) -- (mul.west);
\path [draw, -latex'] (mul.east) -- (af.west);
\path [draw, -latex'] (s.south) |- (st.west);
\path [draw, -latex'] (st.east) -- (ts.west);
\path [draw, -latex'] (ts.east) -| (mul.south);
\end{tikzpicture}

}
  \caption{The whole pipeline}
  \label{fig:pipeline}
\end{figure}
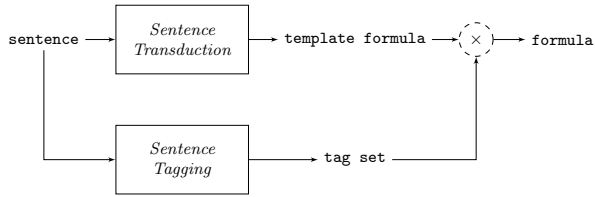

We intend to exploit this behaviour and split the translation of a sentence into a corresponding OWL formula, say \eqref{eq:firstsentence} into \eqref{eq:definition}, into two parallel phases, as depicted in Figure \ref{fig:pipeline}. The \emph{sentence transduction} phase identifies the logical structure of the corresponding formula, returning what we call a \emph{formula template}; for instance, for \eqref{eq:firstsentence}, \eqref{eq:altsentence}, and \eqref{eq:diffsentence} the formula template is:
\begin{equation}
  \label{eq:formula}
  \mathtt{C_0} \sqsubseteq \mathtt{C_1} \sqcap =\mathtt{N_0} \ \mathtt{R_0}.\mathtt{C_2} \sqcap \exists \mathtt{R_1}.\mathtt{C_3}
\end{equation}
where \texttt{C}, \texttt{R} and \texttt{N} have been respectively used for concepts, roles and numbers, with a subscripts indicating their order. Such phase, aims to identify common underlying structural patterns among different text surfaces that can differ both lexically and syntactically.

The \emph{sentence tagging} phase identifies how words act in the sentence: concept (\texttt{C}), role (\texttt{R}), number (\texttt{N}) or any other kind of word (\texttt{w}). The output of this phase can be represented by the following \emph{tagged sequence}:

\begin{align}
  \label{eq:taggedseq}
  \begin{split}
  A \  & [bee]_{\mathtt{C_0}} \ is \ an \ [insect]_{\mathtt{C_1}} \ that \ [has]_{\mathtt{R_0}}
   [6]_{\mathtt{N_0}} \ [legs]_{\mathtt{C_2}} \ and \\ & [produces]_{\mathtt{R_1}} \ [honey]_{\mathtt{C_3}}
  \end{split}
\end{align}

The final step in the pipeline is to combine the outputs of the two phases to obtain the resulting logical formula. Thus, the formula template \eqref{eq:formula} combined with the tagged sentence \eqref{eq:taggedseq} and will provide formula \eqref{eq:definition} the logical conversion for both sentences \eqref{eq:firstsentence}. In this perspective, the ontology learning process can be seen as a \emph{transduction} task, in which a sequence of symbols, a sentence, is translated into another sequence of symbols, a logical formula.

\section{Networks Description}
\label{sec:rnnsforol}

In this section we will present the technical details of the network architectures used to assembly the pipeline depicted in Figure~\ref{fig:pipeline}. Before diving into the network architectures descritpion, we will briefly clarify the notation that will be used henceforth. We will start describing the simplest network architecture, used for the \emph{Sentence Tagging}, implemented through a single Recurrent Neural Network (see ~\cite{mesnil2013investigation}). Then we will describe the system used for the \emph{Sentence Transduction} task, a Recurrent Encoder-Decoder model similar to the one used in \cite{cho2014learning}, wich can be seen as a sort of extension of the previous one. Finally, we will describe the cell model which is used by both the neural networks, namely the Gated Recursive Unit (GRU) model.

\subsection{Notation and conventions.}
We indicate vectors with bold lowercase letters, e.g. $\mathbf{x}$, and matrices with bold uppercase letters, e.g. $\mathbf{W}$. 
Writing explicitly a vector in its components, we use square brackets, like in $\mathbf{x} = [x_1 ... x_n]$. 
To indicate the $j$-th component of the vector, we use the subscript, so that $\mathbf{x}_j = x_j$.
When a vector is made of the concatenation of other vectors, we will write such vectors between square brackets but separated with a semicolon, as in $\mathbf{x} = [\mathbf{x}_1 ; \mathbf{x}_2 ; \mathbf{x}_3]$. 
Uppercase letters will be used to represents sets other than $\mathbb{R}$ for real numbers and $\mathbb{N}$ for nartural numbers. 
Given a set $V$, we will indicate the number of its element with $\vert V \vert$. 
We will use the greek letter $\theta$ to indicate set of trainable parameters of the model. 
Both for vectors and scalars, we indicate the time-step of our model with a superscript between angle brackets, tipically referring to the time-step with the letter $k$, as in $\mathbf{x}^{\langle k \rangle}$ or $x^{\langle k \rangle}$. When a vector represents a time sequence, we will use such notation between square brackets, like in $\mathbf{x} = [x^{\langle 1 \rangle}...x^{\langle n \rangle}]$.

\subsection{Network Model for Sentence Tagging.}
\label{sec:tag-rnn}
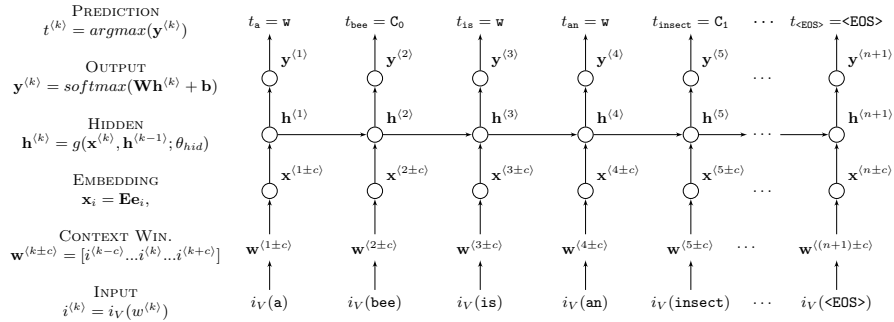
\begin{figure}[t]
  \centering
  \resizebox{\textwidth}{!}{
  \begin{tikzpicture}[node distance=6.5em]

\node [name=input] {%
  \begin{minipage}{0.4\textwidth}
    \centering
    \textsc{Input} \\ 
    $i^{\langle k \rangle} = i_V(w^{\langle k \rangle})$
  \end{minipage}
};
\node [name=a, right of=input, xshift=3em] {$i_{V}(\texttt{a})$};
\node [name=bee, right of=a] {$i_{V}(\texttt{bee})$};
\node [name=is, right of=bee] {$i_{V}(\texttt{is})$};
\node [name=an, right of=is] {$i_{V}(\texttt{an})$};
\node [name=insect, right of=an] {$i_{V}(\texttt{insect})$};
\node [name=dots, right of=insect, xshift=-2em] {\ldots};
\node [name=eos, right of=dots, xshift=-2em] {$i_{V}(\texttt{<EOS>})$};

\node [name=context, above of=input, yshift=-3em] {
\begin{minipage}{0.4\textwidth}
\centering
\textsc{Context Win.} \\
$\mathbf{w}^{\langle k \pm c \rangle} = [i^{\langle k-c \rangle}...i^{\langle k \rangle}...i^{\langle k+c \rangle}]$
\end{minipage}
};
\node [name=w0, right of=context, xshift=3em] {$\mathbf{w}^{\langle 1 \pm c \rangle}$};  
\node [name=w1, right of=w0] {$\mathbf{w}^{\langle 2 \pm c \rangle}$};  
\node [name=w2, right of=w1] {$\mathbf{w}^{\langle 3 \pm c \rangle}$};  
\node [name=w3, right of=w2] {$\mathbf{w}^{\langle 4 \pm c \rangle}$};  
\node [name=w4, right of=w3] {$\mathbf{w}^{\langle 5 \pm c \rangle}$};  
\node [name=wdots, right of=w4, xshift=-3em] {\ldots};  
\node [name=wL, right of=wdots, xshift=-1em] {$\mathbf{w}^{\langle (n+1) \pm c \rangle}$};  

\path [draw, -latex'] (a.north) -- (w0.south);
\path [draw, -latex'] (bee.north) -- (w1.south);
\path [draw, -latex'] (is.north) -- (w2.south);
\path [draw, -latex'] (an.north) -- (w3.south);
\path [draw, -latex'] (insect.north) -- (w4.south);
\path [draw, -latex'] (eos.north) -- (wL.south);

\node [name=embedding, above of=context, yshift=-3em] {
\begin{minipage}{0.4\textwidth}
\centering
\textsc{Embedding} \\
$\mathbf{x}_i = \mathbf{E} \mathbf{e}_i,$
\end{minipage}
};
\node [draw, shape=circle, name=x0, right of=embedding, xshift=3em, label=above right:$\mathbf{x}^{\langle 1 \pm c \rangle}$] {};
\node [draw, shape=circle, name=x1, right of=x0, label=above right:$\mathbf{x}^{\langle 2 \pm c \rangle}$] {};
\node [draw, shape=circle, name=x2, right of=x1, label=above right:$\mathbf{x}^{\langle 3 \pm c \rangle}$] {};
\node [draw, shape=circle, name=x3, right of=x2, label=above right:$\mathbf{x}^{\langle 4 \pm c \rangle}$] {};
\node [draw, shape=circle, name=x4, right of=x3, label=above right:$\mathbf{x}^{\langle 5 \pm c \rangle}$] {};
\node [name=xdots, right of=x4, xshift=-2em] {\ldots};
\node [draw, shape=circle, name=xL, right of=xdots, xshift=-2em, label=above right: {$\mathbf{x}^{\langle n \pm c \rangle}$}] {};

\path [draw, -latex'] (w0.north) -- (x0.south);
\path [draw, -latex'] (w1.north) -- (x1.south);
\path [draw, -latex'] (w2.north) -- (x2.south);
\path [draw, -latex'] (w3.north) -- (x3.south);
\path [draw, -latex'] (w4.north) -- (x4.south);
\path [draw, -latex'] (wL.north) -- (xL.south);

\node [name=recurrent, above of=embedding, yshift=-3em] {
\begin{minipage}{0.4\textwidth}
\centering
\textsc{Hidden} \\
$\mathbf{h}^{\langle k \rangle} = g(\mathbf{x}^{\langle k \rangle}, \mathbf{h}^{\langle k-1 \rangle}; \theta_{hid})$
\end{minipage}
};
\node [draw, shape=circle, name=h0, right of=recurrent, xshift=3em, label=above right:{$\mathbf{h}^{\langle 1 \rangle}$}] {};
\node [draw, shape=circle, name=h1, right of=h0, label=above right:{$\mathbf{h}^{\langle 2 \rangle}$}] {};
\node [draw, shape=circle, name=h2, right of=h1, label=above right:{$\mathbf{h}^{\langle 3 \rangle}$}] {};
\node [draw, shape=circle, name=h3, right of=h2, label=above right:{$\mathbf{h}^{\langle 4 \rangle}$}] {};
\node [draw, shape=circle, name=h4, right of=h3, label=above right:{$\mathbf{h}^{\langle 5 \rangle}$}] {};
\node [name=hdots, right of=h4, xshift=-2em] {\ldots};
\node [draw, shape=circle, name=hL, right of=hdots, xshift=-2em, label=above right:{$\mathbf{h}^{\langle n+1 \rangle}$}] {};

\path [draw, -latex'] (x0.north) -- (h0.south);
\path [draw, -latex'] (x1.north) -- (h1.south);
\path [draw, -latex'] (x2.north) -- (h2.south);
\path [draw, -latex'] (x3.north) -- (h3.south);
\path [draw, -latex'] (x4.north) -- (h4.south);
\path [draw, -latex'] (xL.north) -- (hL.south);

\path [draw, -latex'] (h0.east) -- (h1.west);
\path [draw, -latex'] (h1.east) -- (h2.west);
\path [draw, -latex'] (h2.east) -- (h3.west);
\path [draw, -latex'] (h3.east) -- (h4.west);
\path [draw, -latex'] (h4.east) -- (hdots.west);
\path [draw, -latex'] (hdots.east) -- (hL.west);

\node [name=outputs, above of=recurrent, yshift=-3em] {
\begin{minipage}{0.4\textwidth}
\centering
\textsc{Output} \\
$ \mathbf{y}^{\langle k \rangle} = softmax(\mathbf{W} \mathbf{h}^{\langle k \rangle} + \mathbf{b}) $
\end{minipage}
};
\node [draw, shape=circle, name=y0, right of=outputs, xshift=3em, label=above right:{$\mathbf{y}^{\langle 1 \rangle}$}] {};
\node [draw, shape=circle, name=y1, right of=y0, label=above right:{$\mathbf{y}^{\langle 2 \rangle}$}] {};
\node [draw, shape=circle, name=y2, right of=y1, label=above right:{$\mathbf{y}^{\langle 3 \rangle}$}] {};
\node [draw, shape=circle, name=y3, right of=y2, label=above right:{$\mathbf{y}^{\langle 4 \rangle}$}] {};
\node [draw, shape=circle, name=y4, right of=y3, label=above right:{$\mathbf{y}^{\langle 5 \rangle}$}] {};
\node [name=ydots, right of=y4, xshift=-2em] {\ldots};
\node [draw, shape=circle, name=yL, right of=ydots, xshift=-2em, label=above right:{$\mathbf{y}^{\langle n+1 \rangle}$}] {};

\path [draw, -latex'] (h0.north) -- (y0.south);
\path [draw, -latex'] (h1.north) -- (y1.south);
\path [draw, -latex'] (h2.north) -- (y2.south);
\path [draw, -latex'] (h3.north) -- (y3.south);
\path [draw, -latex'] (h4.north) -- (y4.south);
\path [draw, -latex'] (hL.north) -- (yL.south);

\node [name=tags, above of=outputs, yshift=-3em] {
\begin{minipage}{0.4\textwidth}
\centering
\textsc{Prediction} \\
$t^{\langle k \rangle} = argmax(\mathbf{y}^{\langle k \rangle})$
\end{minipage}
};
\node [name=t_a, right of=tags, xshift=3em] {$t_{\texttt{a}} = \mathtt{w}$};
\node [name=t_bee, right of=t_a] {$t_{\texttt{bee}} = \mathtt{C_0}$};
\node [name=t_is, right of=t_bee] {$t_{\texttt{is}} = \mathtt{w}$};
\node [name=t_an, right of=t_is] {$t_{\texttt{an}} = \mathtt{w}$};
\node [name=t_insect, right of=t_an] {$t_{\texttt{insect}} = \mathtt{C_1}$};
\node [name=t_dots, right of=t_insect, xshift=-2em] {\texttt{\ldots}};
\node [name=t_eos, right of=t_dots, xshift=-2em] {$t_{\texttt{<EOS>}} =$\texttt{<EOS>}};

\path [draw, -latex'] (y0.north) -- (t_a.south);
\path [draw, -latex'] (y1.north) -- (t_bee.south);
\path [draw, -latex'] (y2.north) -- (t_is.south);
\path [draw, -latex'] (y3.north) -- (t_an.south);
\path [draw, -latex'] (y4.north) -- (t_insect.south);
\path [draw, -latex'] (yL.north) -- (t_eos.south);

\end{tikzpicture}

  }%
  \caption{The RNN used for sentence tagging}
  \label{fig:slotfill}
\end{figure}

At a glance, our \emph{Sentence Tagging} task can be formulated as follows: given a natural language sentence corresponding to a DL formula, we want to apply a tag to each word that identifies the role the word has in a the formula. Back to our example in Sec.~\ref{sec:taskdesc}, we want to generate the tags that are applied to the word of our description of a Bee as in ~\ref{eq:taggedseq}. We propose to use a single Recurrent Neural Network with Gated Recursive Units, as synthetically depicted in Fig.~\ref{fig:slotfill}. Such model will be described in detail below, starting from the bottom-most layer of the picture and moving upwards.

\subsubsection{Input Layer.}
\label{sec:tag-rnn-input}
The actual input of our system will be a sentence in natural language. We represent a sentence as a sequence of $n + 1$ symbols, where the first $n$ symbols are the actual words in the sentence and the last one is a conventional symbol, \texttt{<EOS>}, indicating the end of the sequence:
\begin{equation}
\label{eq:sentence}
\mathbf{s} = [w^{\langle 1 \rangle} \, w^{\langle 2 \rangle}...w^{\langle n \rangle} \, w^{\langle n+1 \rangle} = \text{\texttt{<EOS>}}].
\end{equation}
where $w^{\langle k \rangle} \in V, k=1,...,n+1$, being $V$ the \emph{vocabulary}, i.e. the set of all the words that can be used in our language, plus the \texttt{<EOS>} symbol. Each word can be mapped to an integer value, $i_{V}(w)$, representing its \emph{index}, i.e. its position within the vocabulary. As a consequence, each sentence can be represented as an \emph{input sequence} of such indexes:

\begin{equation}
\label{eq:input-sequence}
\mathbf{i}_s = [i_{V}(w^{\langle 1 \rangle}) \, i_{V}(w^{\langle 2 \rangle})...i_{V}(w^{\langle n \rangle}) \, i_{V}(w^{\langle n+1 \rangle}) = i_{V}(\text{\texttt{<EOS>}})].
\end{equation}

At the $k$-th step, the $k$-th element of the input sequence will represent the index of the $k$-th word in the sentence. From now on, we will omit the vocabulary in the subscript and the argument $w$, using the simpler form $i^{\langle k \rangle}$.

\subsubsection{Context Windowing.}
\label{sec:tag-rnn-window}
For each word in the sentence, we consider a window of $c$ words at the left and $c$ words at the right, in order to keep in account the local context in which the word appears. The width of the context is an hyperparameter that has to be set at design time. The $k$-th integer in the input sequence is mapped into a vector $\mathbf{w}^{\langle k \pm c \rangle} \in \mathbb{N}^{(2c) + 1}$ so that:

\begin{equation}
  \label{eq:context-win}
  \mathbf{w}^{\langle k \pm c \rangle} = [i^{\langle k-c \rangle} i^{\langle k-c+1 \rangle}...i^{\langle k \rangle}...i^{\langle k+c-1 \rangle} i^{\langle k+c \rangle}].
\end{equation}

When the context window falls outside the actual boundaries of the sentence, we pad it using the \texttt{<EOS>} symbol. The sequence of such windows is then fed to the next layer of the network.

\subsubsection{Embedding Layer.}
\label{sec:tag-rnn-emb}
Word embedding into a distributed representation is a key concept in using neural networks for Natural Language Processing task. Representing a word $w$ with its position $i(w)$ within a vocabulary $V$ is somehow equivalent to represent it with a one-hot vector $\mathbf{e}_i \in \mathbb{N}^{\vert V \vert}$, i.e. a vector of the same dimension of the vocabulary with the $i$-th component set to 1 and all the others set to 0. So, assuming a vocabulary of 10 words, the 3-rd word will be represented by the one-hot vector:
\begin{equation}
  \label{eq:one-hot}
  \mathbf{e}_3 = [0\,0\,\mathbf{1}\,0\,0\,0\,0\,0\,0\,0]
\end{equation}
Such vectors can be inpractical for several reasons. First of all, they are \emph{large}, as they have the same dimension of the vocabulary which can hold hundreds of thousends words. Second, it can be hard to compose them in a meaningful way. As an example, consider that a dot product between two one-hot vectors will end up only in 1 if the two vectors are the same, 0 otherwise. 

Such problems can be overcome using a \emph{distributed representation}: each word in the vocabulary is associated with a vector of dimension $d$ of real numbers, representing the word in some latent feature space. The dimension of such space $d$ is an hyperparameter and it is tipically way smaller than $\vert V \vert$, helping us to avoid the so called \emph{curse of dimensionality} (see\cite{bengio2003neural}). Moreover, the representations that we obtain in this way are capable to capture regularities in large text corpora, as in \cite{mikolov2013linguistic}. Finally, in recent years, many attempt have been made in using linear algebra in order to compose such vectors in a functional way, ending up having  distributed representations of larger chunk of text -- sentences, paragraphs and entire documents (see \cite{mikolov2013distributed, le2014distributed}). 

The embedding layer of our neural network translate a representation of a word from a single integer -- or a vector of integers, for a context window -- into a distributed one. The intuition is extremely simple: the $i$-th word of the dictionary is associated with the $i$-th column of a $d \times \vert V \vert$ matrix $\mathbf{E}$ of real numbers, representing the distributed representation of the given word, $\mathbf{x}_i$. Operationally, this is equivalent to a multiplication between such matrix and the one-hot column vector $\mathbf{e}_i$ associated to the word:
\begin{equation}
  \label{eq:embedding}
  \mathbf{x}_i = \mathbf{E} \mathbf{e}_i,
\end{equation}
where the subscript $i$ refers to a position within the vocabulary V. The dimensions of the involved matrix are: $\mathbf{x}_i \in \mathbb{R}^{d}$, $\mathbf{E} \in \mathbb{R}^{d \times \vert V \vert}$ and $\mathbf{e}_i \in \mathbb{R}^{\vert V \vert}$.

Our embedding layer is fed at the $k$-th step with a vector of $(2 \times c) + 1$ integers, representing the context window of the $k$-th word. The corresponding distributed representation will be just the \emph{concatenation} of the distributed representation of each word in the context window, namely:
\begin{equation}
  \label{eq:embedding-win}
  \mathbf{x}^{\langle k  \pm c \rangle} = [\mathbf{x}^{\langle k-c \rangle} ; ... \,; \mathbf{x}^{\langle k \rangle} ; ... \,; \mathbf{x}^{\langle k+c \rangle}]
\end{equation}
which is a vector of real numbers of dimension $((2 \times c) + 1) \times d$, being $(2 \times c) + 1$ the number of words in the window and number of embedding vectors to be concatenated and $d$ the dimension of each embedding vector.

The embedding matrix $\mathbf{E}$ can be, like in our case, among the parametres of the model that are learnt during the training phase: in this scenario the model learns also the distributed representation for the words in $V$. We indicate the parameter set for the embeddingg layer as $\theta_{emb} = [ \mathbf{E} ]$. In general, it is also possible to exploit the same pre-trained embeddings and plug them into different models: this can lead to significant reduction of the training phase.

\subsubsection{Recurrent Hidden Layer.}
\label{sec:tag-rnn-hid}
In the so called \emph{deep networks}, the hidden layers are meant to encode the embedded input, capturing some of its features that will be used in the upper layers of the network to predict the output. Dealing with natural language, where each word is bound to other words through syntactic dependencies, such feature extraction phase must keep in account, at each step, the representation of the previous steps. In the recurrent layer, the hidden state at step $k$, $\mathbf{h}^{\langle k \rangle}$ is a function of the $k$-th input step $\mathbf{x}^{\langle k \rangle}$ and of the $(k-1)$-th hidden state, $\mathbf{h}^{\langle k-1 \rangle}$. We can generically write:
\begin{equation}
  \label{eq:recurrent}
  \mathbf{h}^{\langle k \rangle} = g(\mathbf{x}^{\langle k \rangle}, \mathbf{h}^{\langle k-1 \rangle}; \theta_{hid}),
\end{equation}
where $g(\cdot)$ is a generic parametric function of the current input and the previous hidden state and $\theta_{hid}$ is the set of its parameters to be learnt during the training phase. The function that we used is the so called Gated Recursive Unit model and will be described in detail later in ~\ref{sec:gru}. The dimension of the vector $\mathbf{h}^{\langle \cdot \rangle}$, say $h$, is an hyperparameter to be chosen at design time. Such vectors represent the so called \emph{hidden state} of the networks. The hidden vector at step $0$, which is supposed to be the previous hidden state of the actual first one, is initialized to the zero vector.

\subsubsection{Output Layer.}
\label{sec:tag-rnn-out}
As for the vocabulary $V$, we can collect all the possible tags in a catalogue $T$. The output layer is fed with the hidden state of the network and is meant to output, at the $k$-th step, an estimation of the probability of each tag to be the right one for the $k$-th word. Such estimation can be described in different steps. First, the so called \emph{activation} is evaluated though the:
\begin{equation}
  \label{eq:output-activation}
  \mathbf{o}^{\langle k \rangle} = \mathbf{W} \mathbf{h}^{\langle k \rangle} + \mathbf{b},
\end{equation}
where $\mathbf{W}$ is a $\vert T \vert \times h$ matrix and $\mathbf{b}$ is a bias vector of dimension $\vert T \vert$ and both of them are learnt during the training phase, so that, for the output layer, we can write $\theta_{out} = [\mathbf{W}, \mathbf{b}]$.

The activation vector $\mathbf{o}^{\langle k \rangle}$ is a vector of the same dimension of the output vocabulary, as already said, but it can't yet describe a probability distribution. We apply a $softmax(\cdot)$ function to such vector in order to obtain the actual output vector:
\begin{equation}
  \label{eq:output-softmax}
  \mathbf{y}^{\langle k \rangle} = softmax(\mathbf{o}^{\langle k \rangle}).
\end{equation}
The $i$-th component of $\mathbf{y}^{\langle k \rangle}$ will now be given by the softmax function:
\begin{equation}
  \label{eq:softmax}
  {y}_i^{\langle k \rangle} = \frac{
    exp({o}_i^{\langle k \rangle})
  }
  {
    \sum_{j=1}^{\vert T \vert}{exp({o}_j^{\langle k \rangle})}
  },
\end{equation}
so that all the components of the $\mathbf{y}^{\langle k \rangle}$ vector sum up to one: in this way, the $i$-th component of the vector can model the probability of the $i$-th tag in the tag catalogue $T$ to be the correct tag for the $k$-th word. More in detail, our model estimates the probability of having a certain tag for the $k$-th word given the word itself and all the previous ones -- with their context, omitted in the notation -- and the parameters of the model:
\begin{equation}
  \label{eq:tagging-prob}
  \mathbf{y}^{\langle k \rangle} = P(t^{\langle k \rangle}|w^{\langle k \rangle}, w^{\langle k-1 \rangle},...\,, w^{\langle 0 \rangle}; \theta),
\end{equation}
where $\theta = [\theta_{emb}, \theta_{hid}, \theta_{out}]$ is the collection of all the parameters used in the various layers.

\subsubsection{Prediction.}
\label{sec:tag-rnn-pred}
The predicted tag for the $k$-th word will be the one with the highest probability as modeled by the $\mathbf{y}^{\langle k \rangle}$ vector. Applying the $argmax(\cdot)$  to such vector we will obtain an integer value
\begin{equation}
  \label{eq:prediction}
  t^{\langle k \rangle} = argmax(\mathbf{y}^{\langle k \rangle}).
\end{equation}
indicating the position of the predicted tag in $T$.

\subsubsection{Loss Function and Training Objective.}
\label{sec:tag-rnn-trainobj}
Given a training set of tagged sentences $S$ of length $l$\footnote{It means that all the sentences will be padded to the same length adding trailing \texttt{<EOS>} symbols to the shorter ones.}, training such model will end up in minimizing the sum of the categorical cross entropy between the predicted tag sequence and the gold tag sequence. In details, let $(\mathbf{s}, \mathbf{t})$ be a training example; $s$ is a sentence as in (\ref{eq:sentence}) and a $\mathbf{t}$ is a tag sequence $\mathbf{t} = [t^{\langle 0 \rangle}t^{\langle 1 \rangle}...t^{\langle n+1 \rangle}=\text{\texttt{<EOS>}}]$. The $k$-th element in the tag sequence, $t^{\langle k \rangle}$, is a integer value representing the position within the tag set $T$ of the tag to be applied to the $k$-th word in the sentence. As for the words in the input layer, we can represent such value with a one-hot vector of dimension $\vert T \vert$ with all elements set to 0 but the $t^{\langle k \rangle}$-th, which is set to 1. Let $\mathbf{\hat{y}}^{\langle k \rangle}$ be such vector. The difference between the predicted tag sequence and the expected one, can be measured through the categorical cross entropy:
\begin{equation}
  \label{eq:cce-tags}
  \mathcal{L}_{s} = - \sum_{k=1}^{n+1}{\mathbf{\hat{y}}^{\langle k \rangle} log(\mathbf{y}^{\langle k \rangle})},
\end{equation}
and being $\mathbf{\hat{y}}^{\langle k \rangle}$ a one-hot vector, the dot product between such vector and $\mathbf{y}^{\langle k \rangle}$ will end up in the product between it's $t^{\langle k \rangle}$-th component, the only non-zero one, and the log of corresponding component of the output vector $\mathbf{y}^{\langle k \rangle}$, namely 
\begin{equation}
  \label{eq:cce-tags-2}
  \mathcal{L}_{s} = - \sum_{k=1}^{n+1}{\mathbf{\hat{y}}^{\langle k \rangle}_{t^{\langle k \rangle}} log(\mathbf{y}^{\langle k \rangle}_{t^{\langle k \rangle}})}.
\end{equation}
Moreover, being again $\mathbf{\hat{y}}^{\langle k \rangle}$ a one-hot vector, its only non-zero component $\mathbf{\hat{y}}^{\langle k \rangle}_{t^{\langle k \rangle}}$ will have value 1, so the previous equation can be simplified to:
\begin{equation}
  \label{eq:cc}
  \mathcal{L}_{s} = - \sum_{k=1}^{n+1}{log(\mathbf{y}^{\langle k \rangle}_{t^{\langle k \rangle}})}.
\end{equation}
The argument of the logarithm is a real number between 0 and 1, so each term of the summation will be non-positive; the minus sign in front of the summation ensures that the loss term for the sentence $\mathcal{L}_s$ is always non-negative. Finally, being $\vert S \vert$ the number of examples in our training set $S$, the global loss will be the sum of the loss of each sentence, namely:
\begin{equation}
  \label{eq:loss}
  \mathcal{L} = \sum_{s \in S}\mathcal{L}_{s},
\end{equation}
which is non-negative, as a summation of non-negative terms. Training the network architecture described so far means basically to minimize such loss value w.r.t the parameter set $\theta$.

\subsection{Network Model for Sentence Transduction}
\label{sec:trans-red}

\begin{figure}[tbh]
  \centering
    \resizebox{\textwidth}{!}{
  \begin{tikzpicture}[node distance=6em]

\node [name=input] {
  \begin{minipage}{0.4\textwidth}
    \centering
    \textsc{Input} \\ 
    $i^{\langle k \rangle} = i_V(w^{\langle k \rangle})$
  \end{minipage}
};
\node [name=a, right of=input, xshift=2em] {$i_{\texttt{a}}$};
\node [name=bee, right of=a] {$i_{\texttt{bee}}$};
\node [name=is, right of=bee] {$i_{\texttt{is}}$};
\node [name=an, right of=is] {$i_{\texttt{an}}$};
\node [name=insect, right of=an] {$i_{\texttt{insect}}$};
\node [name=dots, right of=insect, xshift=-1em] {\texttt{\ldots}};
\node [name=eos, right of=dots, xshift=-1em] {$i_{\texttt{<EOS>}}$};

\node [name=embedding, above of=input, yshift=-2em] {
\begin{minipage}{0.4\textwidth}
\centering
\textsc{Embedding} \\
$\mathbf{x}_i = \mathbf{E} \mathbf{e}_i,$
\end{minipage}
};
\node [draw, shape=circle, name=x0, right of=embedding, xshift=2em, label=above right:$\mathbf{x}^{\langle 1 \rangle}$] {};
\node [draw, shape=circle, name=x1, right of=x0, label=above right:$\mathbf{x}^{\langle 2 \rangle}$] {};
\node [draw, shape=circle, name=x2, right of=x1, label=above right:$\mathbf{x}^{\langle 3 \rangle}$] {};
\node [draw, shape=circle, name=x3, right of=x2, label=above right:$\mathbf{x}^{\langle 4 \rangle}$] {};
\node [draw, shape=circle, name=x4, right of=x3, label=above right:$\mathbf{x}^{\langle 5 \rangle}$] {};
\node [name=xdots, right of=x4, xshift=-1em] {\ldots};
\node [draw, shape=circle, name=xL, right of=xdots, xshift=-1em, label=above right: {$\mathbf{x}^{\langle n + 1 \rangle} =$ \texttt{<EOS>}}] {};

\path [draw, -latex'] (a.north) -- (x0.south);
\path [draw, -latex'] (bee.north) -- (x1.south);
\path [draw, -latex'] (is.north) -- (x2.south);
\path [draw, -latex'] (an.north) -- (x3.south);
\path [draw, -latex'] (insect.north) -- (x4.south);
\path [draw, -latex'] (eos.north) -- (xL.south);

\node [name=encoding, above of=embedding, yshift=-2em] {
\begin{minipage}{0.4\textwidth}
\centering
\textsc{Encoding} \\
$\mathbf{h}_{e}^{\langle k \rangle} = g_{e}(\mathbf{x}^{\langle k \rangle}, \mathbf{h}^{\langle k-1 \rangle}; \theta_{enc})$
\end{minipage}
};
\node [draw, shape=circle, name=h0, right of=encoding, xshift=2em, label=above right:{$\mathbf{h}_e^{\langle 1 \rangle}$}] {};
\node [draw, shape=circle, name=h1, right of=h0, label=above right:{$\mathbf{h}_e^{\langle 2 \rangle}$}] {};
\node [draw, shape=circle, name=h2, right of=h1, label=above right:{$\mathbf{h}_e^{\langle 3 \rangle}$}] {};
\node [draw, shape=circle, name=h3, right of=h2, label=above right:{$\mathbf{h}_e^{\langle 4 \rangle}$}] {};
\node [draw, shape=circle, name=h4, right of=h3, label=above right:{$\mathbf{h}_e^{\langle 5 \rangle}$}] {};
\node [name=hdots, right of=h4, xshift=-1em] {\ldots};
\node [draw, shape=circle, name=hL, right of=hdots, xshift=-1em, label=above right:{$\mathbf{h}_e^{\langle n + 1 \rangle} = \mathbf{c}$}] {};

\path [draw, -latex'] (x0.north) -- (h0.south);
\path [draw, -latex'] (x1.north) -- (h1.south);
\path [draw, -latex'] (x2.north) -- (h2.south);
\path [draw, -latex'] (x3.north) -- (h3.south);
\path [draw, -latex'] (x4.north) -- (h4.south);
\path [draw, -latex'] (xL.north) -- (hL.south);

\path [draw, -latex'] (h0.east) -- (h1.west);
\path [draw, -latex'] (h1.east) -- (h2.west);
\path [draw, -latex'] (h2.east) -- (h3.west);
\path [draw, -latex'] (h3.east) -- (h4.west);
\path [draw, -latex'] (h4.east) -- (hdots.west);
\path [draw, -latex'] (hdots.east) -- (hL.west);

\node [name=decoding, above of=encoding, yshift=-1em] {
\begin{minipage}{0.4\textwidth}
\centering
\textsc{Decoding} \\
$\mathbf{h}_{d}^{\langle k \rangle} = g_{d}(\mathbf{x}^{\langle k \rangle}, \mathbf{h}^{\langle k-1 \rangle}; \theta_{dec})$
\end{minipage}
};
\node [draw, shape=circle, name=h_d0, right of=decoding, xshift=2em, label=above right:{$\mathbf{h}_d^{\langle 1 \rangle}$}] {}; 
\node [draw, shape=circle, name=h_d1, right of=h_d0, label=above right:{$\mathbf{h}_d^{\langle 2 \rangle}$}] {}; 
\node [draw, shape=circle, name=h_d2, right of=h_d1, label=above right:{$\mathbf{h}_d^{\langle 3 \rangle}$}] {}; 
\node [name=h_ddots, right of=h_d2, xshift=-1em] {\ldots};
\node [draw, shape=circle, name=h_dL, right of=h_ddots, xshift=-1em, label=above right:{$\mathbf{h}_d^{\langle m + 1 \rangle} = \mathbf{c}$}] {};

\path [draw, dashed, -latex'] (hL.north) -- (h_d0.south);
\path [draw, dashed, -latex'] (hL.north) -- (h_d1.south);
\path [draw, dashed, -latex'] (hL.north) -- (h_d2.south);
\path [draw, dashed, -latex'] (hL.north) -- (h_dL.south);

\path [draw, -latex'] (h_d0.east) -- (h_d1.west);`
\path [draw, -latex'] (h_d1.east) -- (h_d2.west);
\path [draw, -latex'] (h_d2.east) -- (h_ddots.west);
\path [draw, -latex'] (h_ddots.east) -- (h_dL.west);

\node [name=outputs, above of=decoding, yshift=-2em] {
\begin{minipage}{0.4\textwidth}
\centering
\textsc{Output} \\
$ \mathbf{y}^{\langle k \rangle} = softmax(\mathbf{W} \mathbf{h}_d^{\langle k \rangle} + \mathbf{b}) $
\end{minipage}
};
\node [draw, shape=circle, name=y0, right of=outputs, xshift=2em, label=above right:{$\mathbf{y}^{\langle 1 \rangle}$}] {};
\node [draw, shape=circle, name=y1, right of=y0, label=above right:{$\mathbf{y}^{\langle 2 \rangle}$}] {};
\node [draw, shape=circle, name=y2, right of=y1, label=above right:{$\mathbf{y}^{\langle 3 \rangle}$}] {};
\node [name=ydots, right of=y2, xshift=-1em] {\ldots};
\node [draw, shape=circle, name=yL, right of=ydots, xshift=-1em, label=above right:{$\mathbf{y}^{\langle m \rangle}$}] {};

\path [draw, -latex'] (h_d0.north) -- (y0.south);
\path [draw, -latex'] (h_d1.north) -- (y1.south);
\path [draw, -latex'] (h_d2.north) -- (y2.south);
\path [draw, -latex'] (h_dL.north) -- (yL.south);

\node [name=terms, above of=outputs, yshift=-2em] {
\begin{minipage}{0.4\textwidth}
\centering
\textsc{Prediction} \\
$t^{\langle k \rangle} = argmax(\mathbf{y}^{\langle k \rangle})$
\end{minipage}
};
\node [name=t0, right of=terms, xshift=2em] {$\mathtt{C_0}$};
\node [name=t1, right of=t0] {$\sqsubseteq$};
\node [name=t2, right of=t1] {$\mathtt{C_1}$};
\node [name=tdots, right of=t2, xshift=-1em] {\texttt{\ldots}};
\node [name=tL, right of=tdots, xshift=-1em] {\texttt{<EOS>}};

\path [draw, -latex'] (y0.north) -- (t0.south);
\path [draw, -latex'] (y1.north) -- (t1.south);
\path [draw, -latex'] (y2.north) -- (t2.south);
\path [draw, -latex'] (yL.north) -- (tL.south);

\end{tikzpicture}

  }%
  \caption{The RNN Encoder-Decoder network model for sentence transduction}
  \label{fig:network}
\end{figure}
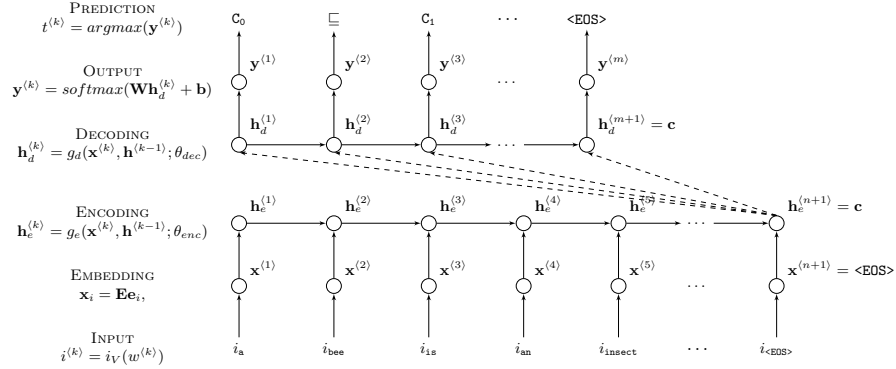

At a glance, our \emph{Sentence Transduction} task can be formulated as follows: given a natural language sentence corresponding to a DL formula, we want to identify the structure of such formula, how its concepts and roles are connected and which connectors are used. Back to our example in Sect.~\ref{sec:taskdesc}, we want to generate the formula template for the description of a Bee as in ~(\ref{eq:formula}). The model we propose for the sentence transduction task is a Recurrent Encoder-Decoder (RE-D), based on the model in \cite{cho2014learning}, but slightly simplified. The \emph{building blocks} are the same components of the model used for the \emph{Sentence Tagging} task described in the previous subsection, so we will avoid technical details when not strictly necessary. The overall architecture is depicted in Fig.~\ref{fig:network} and will be described in details later.
\subsubsection{Input and Embedding Layers.} 
\label{sec:trans-red-input-emb}
The input layer works exactly as the one described in \ref{sec:tag-rnn-input}. In this case, it is not followed by any context windowing phase and the input sequence for each sentence is fed directly into the embedding layer. Also the embedding layer works in the same way as seen before, with the embedding matrix to be learnt at training phase.

\subsubsection{Encoding layer.}
\label{sec:trans-red-enc}
Also the encoding layer works exactly as the one used in the previous model, so that, at the $k$-th step, the hidden state is function of the $k$-th input vector and the $(k-1)$-th hidden state, given the set of parameters $\theta_{enc}$. We can write it as:
\begin{equation}
  \label{eq:trans-red-enc}
  \mathbf{h}_e^{\langle k \rangle} = g_e(\mathbf{x}^{\langle k \rangle}, \mathbf{h}^{\langle k-1 \rangle}; \theta_{enc}),
\end{equation}
with the subscript $e$ standing for \emph{encoding}. 
As for the model used for \emph{Sequence Tagging}, the function $g_e(\cdot)$ is a generic parametric function of the current input and the previous hidden state, while $\theta_{enc}$ is the set of its parameters to be learnt during the training phase. The Gated Recursive Unit model is used as well (see Subsect.~\ref{sec:gru}). The dimension of the vector $\mathbf{h}_e^{\langle \cdot \rangle}$, say $h_e$, is an hyperparameter to be chosen at design time.

In order to better understand the main intuition underlying such architecture, we point out that, due to the recurrent structure, the hidden state at the $k$-th time step is a representation of the sentence so far. So, at the last step, we can think about the state $\mathbf{c} = \mathbf{h}_e^{\langle n+1 \rangle}$ as a representation of the whole sentence. As in \ref{sec:tag-rnn-hid}, the hidden state at step 0, $\mathbf{h}_e^{\langle 0 \rangle}$, is initialized with the zero vector.

\subsubsection{Decoding layer.}
\label{sec:trans-red-dec}
The decoding layer acts as a dual of the encoding one: starting from the vector $\mathbf{c}$ encoding the whole input sentence, a new sequence of hidden states is generated. The $\mathbf{c}$ vector is fed into this layer as a constant input, carrying the meaning of the whole input sentence. This is indeed the main intuition underlying such model: feeding the decoding layer with such \emph{context input} at each step, will allow the encoding and the decoding sequences to be \emph{lousy coupled} w.r.t. their structures but \emph{strongly coupled} w.r.t. their meaning. We can express the hidden decoding state at the $i$-th step as:

\begin{equation}
  \label{eq:trans-red-dec}
  \mathbf{h}_d^{\langle k \rangle} = g_d(\mathbf{c}, \mathbf{h}_d^{\langle k-1 \rangle}; \theta_{dec})
\end{equation}

where the function $g_d(\cdot)$ is a generic parametric function of the constant input $\mathbf{c}$ and the previous hidden state, while $\theta_{dec}$ is the set of its parameters to be learnt during the training phase. The Gated Recursive Unit model is used as well (see Subsect.~\ref{sec:gru}). The dimension of the vector $\mathbf{h}_d^{\langle \cdot \rangle}$, say $h_d$, is an hyperparameter to be chosen at design time. As in \ref{sec:tag-rnn-hid}, the hidden state at step 0, $\mathbf{h}_d^{\langle 0 \rangle}$,is initialized with the zero vector.

\subsubsection{Output Layer, Prediction, Loss Function and Training Objective.}
\label{sec:trans-red-out}
The output layer works exactly as the same for the \emph{Sentence Tagging}, but the output set is given by the set of all the possible formula terms, say $F$. The probability distribution we are modeling now, is the one of having a certain term at the $j$-th position of the formula, $f^{\langle j \rangle}$, given the whole input sentence $s$ and the model parameters $\theta$:
\begin{equation}
  \label{eq:trans-red-prob}
  \mathbf{y}^{\langle j \rangle} = P(f^{\langle j \rangle}|\mathbf{s}; \theta),
\end{equation}
where $\theta = [\theta_e, \theta_{enc}, \theta_{dec}, \theta_{out}]$. We used $j$ instead of $k$ to stress that the input sequence, namely the sentence, and the output one, the formula, \emph{don't have the same length}, while this condition holds in \emph{Sequence Tagging} task. The predicted formula term at the $j$-th step will be the one with the max probability, as in the previous model:
\begin{equation}
  \label{eq:trans-red-pred}
  f^{\langle j \rangle} = argmax(\mathbf{y}^{\langle j \rangle}).
\end{equation}

Said $S = \{ (\mathbf{s}, \mathbf{f})_1,...(\mathbf{s}, \mathbf{f})_{\vert S \vert} \}$ a training set of pairs of sentences and corresponding formulae of length $m + 1$\footnote{As alread said, it means that all the sentences will be padded to the same length adding trailing \texttt{<EOS>} symbols to the shorter ones; the $+1$ term come from the final \texttt{<EOS>} symbol.}, we will use the categorical cross entropy as the loss function. For each sentence in the training set, we will have:
\begin{equation}
  \label{eq:trans-red-loss-ex}
  \mathcal{L} = - \sum_{j=1}^{m+1}{\mathbf{\hat{y}}_i^{\langle j \rangle} log(\mathbf{y}_i^{\langle j \rangle})},
\end{equation}
where $\mathbf{\hat{y}}^{\langle j \rangle}$ is one-hot vector equivalent to the the $j$-th symbol in the gold formula and $\mathbf{y}^{\langle j \rangle}$ is the output vector for the $j$-th step. Summing up such terms for all the examples in the training set, we get the value of the global loss function:
\begin{equation}
  \label{eq:trans-red-loss-ex}
  \mathcal{L} = \sum_{s \in S}{\mathcal{L}_s}.
\end{equation}
All the consideration on the loss function evaluation made in Subsect. ~\ref{sec:tag-rnn-trainobj}, still hold.

\subsection{Gated Recursive Unit.}
\label{sec:gru}
We intend to use the Gated Recursive Unit function in (\ref{eq:recurrent}), (\ref{eq:trans-red-enc}) and (\ref{eq:trans-red-dec}), in order to provide our recurrent neural networks with short-term memory effect. The cell behavior is driven by two gate functions, whose value ranges in the $[0;1]$ interval. The \emph{reset gate} $\mathbf{r}$ and the \emph{update gate} $\mathbf{z}$ which, at the $k$-th step, are defined as follows:

\begin{equation}
  \label{eq:resetgate}
  \mathbf{r}^{\langle k \rangle} = \sigma (\mathbf{W}_r\mathbf{x}^{\langle k \rangle} + \mathbf{U}_r \mathbf{h}^{\langle k-1 \rangle})
\end{equation}

\begin{equation}
    \label{eq:updategate}
  \mathbf{z}^{\langle k \rangle} = \sigma (\mathbf{W}_z\mathbf{x}^{\langle k \rangle} + \mathbf{U}_z \mathbf{h}^{\langle k-1 \rangle})
\end{equation}

where $\mathbf{W}_r$, $\mathbf{U}_r$, $\mathbf{W}_z$, $\mathbf{U}_z$ are weight matrices that are learned during the training, $\mathbf{x}^{\langle k \rangle}$ is the current input of the cell, $\mathbf{h}^{\langle k-1 \rangle}$ is the previous hidden state, and $\sigma$ indicates the element-wise logistic sigmoid function. Said $\odot$ the element-wise product between two vectors, and defined the quantity:

\begin{equation}
  \label{eq-plainactivation}
  \tilde{\mathbf{h}}^{\langle k \rangle} = tanh(\mathbf{W}_h\mathbf{x}^{\langle k \rangle} + \mathbf{r} \odot \mathbf{U}\mathbf{h}^{\langle k-1 \rangle}),
\end{equation}

where $\mathbf{W}$ and $\mathbf{U}$ are learnt weight matrices, the actual activation of the hidden state is given by the function $g$ defined as follows:

\begin{equation}
  \label{eq-hiddenactivation}
  \mathbf{h}^{\langle k \rangle} = g(\mathbf{x}^{\langle k \rangle}, \mathbf{h}^{\langle k - 1 \rangle}) = \mathbf{z} \odot \mathbf{h}^{\langle k-1 \rangle} + (\mathbf{1} - \mathbf{z}) \odot \tilde{\mathbf{h}}^{\langle k \rangle}
\end{equation}

\begin{figure}[t]
  \centering
  \resizebox{0.7\textwidth}{!}{
  \begin{tikzpicture}[node distance=6em]
\node [name=x] {$\mathbf{x}$};
\node [draw, dashed, name=htmp, right of=x, minimum height=2.5em, text width=2.5em, text centered, xshift=-2em] {$\tilde{\mathbf{h}}$};
\node [draw, dashed, shape=circle, right of=htmp, name=r] {$\mathbf{r}$};
\node [draw, name=h, right of=r, minimum height=2.5em, text width=2.5em, text centered] {$\mathbf{h}$};
\node [draw, name=z, above of=h, shape=circle, dashed, yshift=-1.5em] {$\mathbf{z}$};

\path [draw, -latex'] (x.east) -- (htmp.west);
\path [draw, -latex', dashed] (h.west) -- node[below] {{\tiny $\mathbf{h}^{\langle i-1 \rangle}$}} (r.east);
\path [draw, -latex'] (r.west) -- (htmp.east);
\path [draw, -latex'] (z.south) -- (h.north);
\path [draw, -latex'] (htmp.north) |- (z.west);
\path [draw, -latex', dashed] (h.east) -- ++(1em, 0) |- node[below right] {{\tiny $\mathbf{h}^{\langle i-1 \rangle}$}} (z.east);
\end{tikzpicture}

  }%
  \caption{Gated Recursive Unit}
  \label{fig-gru}
\end{figure}
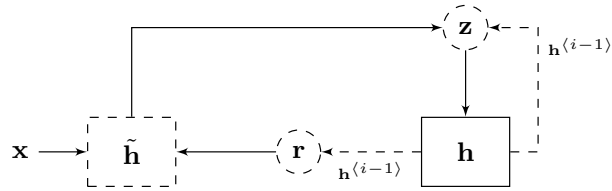

The intuition behind this model is the following. In (\ref{eq-plainactivation}), when the reset gate value gets close to $0$, the new hidden state value ignores its previous value, dropping such piece of information. At the same time, in (\ref{eq-hiddenactivation}), the update gate balances the amount of information to be kept from the previous hidden state and from the current state. 
The set of the parameters to be learnt in the training phase can be summed up as $\theta_h = [\mathbf{W}_r, \mathbf{U}_r, \mathbf{W}_z, \mathbf{U}_z, \mathbf{W}_h, \mathbf{U}_h]$.
The whole cell model can be synthetically depicted as in Fig.~\ref{fig-gru}.

\section{Conclusion and future work}
\label{sec:conclusion}

In this technical report, we described a Recurrent Neural Network based system to be used in an ontology learning process which can be seen as a machine translation task. Such system is currently under evaluation. Our main objective is to verify if such approach can help the ontology engineering community tackling its long term challenges such as achieving some sort of domain independence, reducing the engineering cost and the ability to deal with unconstrained language.


\bibliographystyle{splncs}
\bibliography{references}

\end{document}